\documentclass[sigconf]{acmart}

\usepackage{multirow}
\usepackage{balance}

\AtBeginDocument{%
  \providecommand\BibTeX{{%
    \normalfont B\kern-0.5em{\scshape i\kern-0.25em b}\kern-0.8em\TeX}}}

\copyrightyear{2021}
\acmYear{2021}
\setcopyright{acmcopyright}\acmConference[MM '21]{Proceedings of the 29th ACM
	International Conference on Multimedia}{October 20--24, 2021}{Virtual Event, China}
\acmBooktitle{Proceedings of the 29th ACM International Conference on Multimedia
	(MM '21), October 20--24, 2021, Virtual Event, China}
\acmPrice{15.00}
\acmDOI{10.1145/3474085.3475398}
\acmISBN{978-1-4503-8651-7/21/10}

\begin{document}
\fancyhead{}

\title{Single Image 3D Object Estimation \\ with Primitive Graph Networks}

\author{Qian He$^{1,2,3}$, Desen Zhou$^{4}$, Bo Wan$^{5}$, Xuming He$^{1,6}$}
\authornote{Corresponding author: Xuming He. This work was supported by Shanghai Science and Technology Program 21010502700. }
\affiliation{
	\institution{
		$^{1}$School of Information Science and Technology, ShanghaiTech University\\
		$^{2}$Shanghai Institute of Microsystem and Information Technology, Chinese Academy of Sciences\\
		$^{3}$University of Chinese Academy of Sciences\\
		$^{4}$Department of Computer Vision Technology (VIS), Baidu Inc.\\
		$^{5}$Department of Electrical Engineering (ESAT), KU Leuven\\
		$^{6}$Shanghai Engineering Research Center of Intelligent Vision and Imaging}
	\city{}
	\country{}}
\email{{heqian, wanbo, hexm}@shanghaitech.edu.cn, zhoudesen@baidu.com}

\renewcommand{\shortauthors}{Trovato and Tobin, et al.}

\begin{abstract}
Reconstructing 3D object from a single image (RGB or depth) is a fundamental problem in visual scene understanding and yet remains challenging due to its ill-posed nature and complexity in real-world scenes. To address those challenges, we adopt a primitive-based representation for 3D object, and propose a two-stage graph network for primitive-based 3D object estimation, which consists of a sequential proposal module and a graph reasoning module. Given a 2D image, our proposal module first generates a sequence of 3D primitives from input image with local feature attention. Then the graph reasoning module performs joint reasoning on a primitive graph to capture the global shape context for each primitive. Such a framework is capable of taking into account rich geometry and semantic constraints during 3D structure recovery, producing 3D objects with more coherent structure even under challenging viewing conditions. We train the entire graph neural network in a stage-wise strategy and evaluate it on three benchmarks: Pix3D, ModelNet and NYU Depth V2. Extensive experiments show that our approach outperforms the previous state of the arts with a considerable margin.
\end{abstract}




\maketitle

\section{Introduction}

As a fundamental problem in visual scene understanding, 3D object recovery from a monocular image (RGB or depth) plays an indispensable role in many vision tasks when multi-view sensors are difficult to deploy. 
It has also gained increasing attention in multimedia community~\cite{lu2020single,wang20173densinet,zhu2018learning,meng2020lgnn,zhang2019danet,huang2020hot,zhang2020adaptive,wu2020mm} due to its great potential in VR/AR content generation and robot-scene interaction. 
However, unlike 3D reconstruction in multi-view settings, single-image 3D object estimation remains highly challenging due to the real-world imaging process. A variety of factors, such as (self-)occlusion, varying viewing angles and large variation in object appearance/shapes, lead to a complex mapping from input observation to 3D structure.  

Extensive effort has been made recently to address this problem via learning a deep neural network to map the input image to an output 3D shape. One dominating strategy represents the 3D object shapes in voxel~\cite{wu2016learning,choy20163d}, point cloud~\cite{fan2017point} or mesh~\cite{kato2018neural,Wang_2018_ECCV}. However, such holistic geometric representations are sensitive to noisy inputs and lack the capacity of encoding higher-level regularity in object shapes, such as part symmetry or functionality. 

\begin{figure}[t!]
	\renewcommand\arraystretch{0.005}
	\vspace{-0.0cm}
	\begin{center}
		\scalebox{0.47}[0.47]{
			\begin{tabular}{c}
				\raisebox{-0.9\height}{\includegraphics[width=\textwidth]{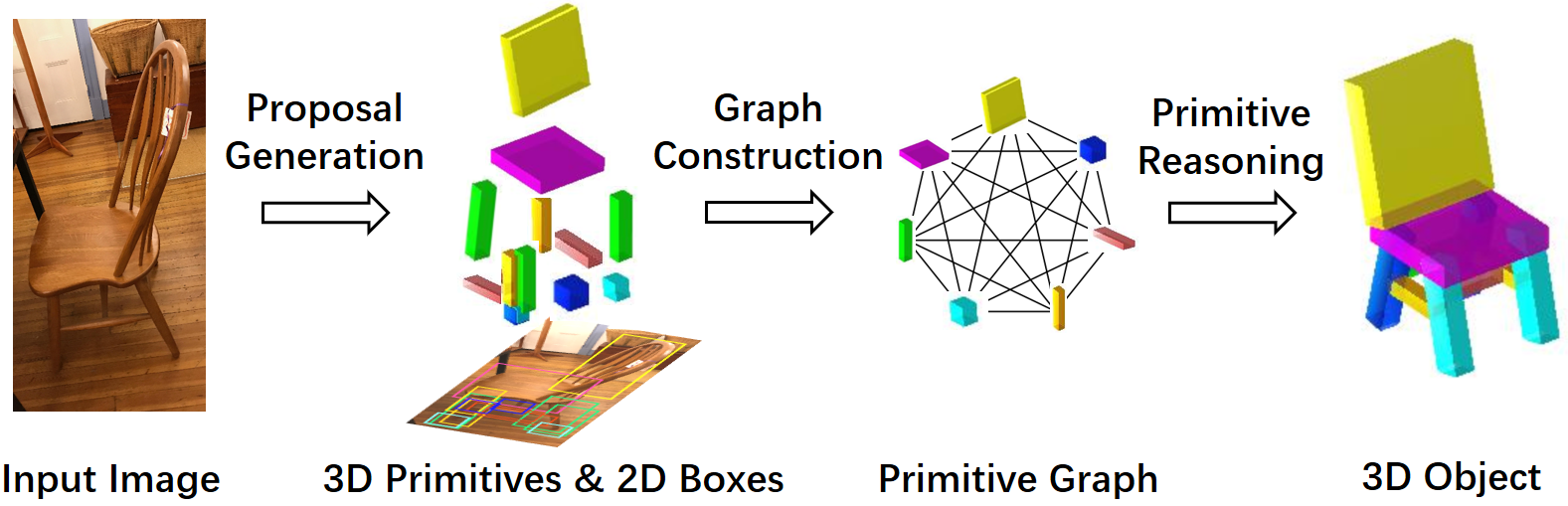}}
			\end{tabular}
		}
	\end{center}
	\caption{Given an input image, our network first generates primitive proposals attending to local features, then builds a graph on the proposals for primitive reasoning, and finally outputs a 3D object shape in primitive representation.}
	\label{fig:adv}
\end{figure}

The primitive-based methods, which exploits the compositionality in object shapes~\cite{biederman1987recognition}, provide an alternative solution that has potential to overcome the above limitations. They typically employ deep networks with an encoder-decoder architecture, which first extract a holistic feature representation of the input and then decode it into a collection of shape primitives. 
For instance, 3D-PRNN~\cite{zou20173d} and PQ-Net~\cite{wu2020pq} predicts a sequence of 3D rectangles, while Im2Struct~\cite{niu2018im2struct} generates a tree-structured representation of object parts with a recursive neural network.  
Most existing primitive-based approaches, however, suffer from several limitations in practice: First, their inference process typically generates primitives in a progressive manner, and hence are prone to error propagation in part estimation. Additionally, their decoders rely on a holistic image representation, which is sensitive to occlusion. Furthermore, part semantics are largely ignored in geometric modeling, often leading to unreasonable 3D object configurations.         

In this paper, we propose a novel primitive-based 3D object recovery framework to address the afore-mentioned limitations. Our main idea is to use a two-stage strategy, in which we first generate a pool of primitive proposals from the input image and then build a fully-connected primitive graph to perform joint reasoning on object configurations (see Fig.~\ref{fig:adv}). Our graph-based primitive inference allows us to better capture the global shape context, mitigate the errors in proposal generation and predict the primitives coherently. Besides, each primitive proposal can attend to its local features and overall the object recovery is less susceptible to occlusion. Finally, with additional semantic part annotations, our method can easily incorporate semantic part relations into the primitive reasoning, which enforces stronger structure constraints for 3D estimation.       


To achieve this, we develop a deep graph neural network that takes a monocular RGB or depth image of an object as input and generate its 3D primitive representation. Our model consists of two main modules, a primitive proposal network and a primitive reasoning network. The proposal module first computes a feature map by a deep Convnet and then adopts a recurrent network to lift the 2D feature into a sequence of 3D rectangular primitive proposals. The reasoning module builds a fully-connected graph network on the proposals, where each node embeds the geometry and semantic features of a primitive and each edge captures the relative geometric and semantic relations. The graph network then refines the primitive features by performing message-passing to capture the global context of object shape. Based on those refined representations, we finally predict the confidence scores (foreground vs. background or part label) and the geometric parameters of all the primitives.            
By exploiting its modular architecture, we train the entire primitive model with a stage-wise strategy, which simplifies the learning procedure and yet works effectively in practice. 

We evaluate our method on three benchmarks: a subset of Pix3D~\cite{sun2018pix3d}, ModelNet~\cite{zou20173d} and NYU Depth V2~\cite{silberman2012indoor}. 
The empirical results and ablation study show that our method consistently outperforms the prior state of the art~\cite{zou20173d,wu2020pq} with a considerable margin. The main contributions of our work are three-folds:
\begin{itemize}
	\item We propose a new primitive-based method for 3D object recovery from a single image, achieving state-of-the-art performance on three benchmarks.
	\item We introduce a modular graph neural network that effectively captures 3D shape constraints and performs joint reasoning on the primitives for coherent 3D estimation.
	\item We enrich the primitive representation by its conv features for robustness in occlusion and optionally by its semantic part cues to enforce stronger constraints on object structure.   
\end{itemize}

\section{Related Work}
\subsection{Single-view 3D Object Estimation}
Recovering 3D objects from single images is an ill-posed problem. Previous works leverage rich CAD models and tackle this problem by joint analysis of images and 3D shapes~\cite{li2015joint,huang2015single}, exploring local correspondence between images and 3D shape~\cite{kong2017using}, or applying deformation from a mean shape~\cite{kar2015category,kanazawa2018learning}.
More recently, extensive approaches for single image 3D object estimation explore various shape representation, such as key-point~\cite{wu2016single,li2017deep,suwajanakorn2018discovery,he20183d}, voxel~\cite{wu2016learning,choy20163d}, point cloud~\cite{fan2017point}, mesh~\cite{kato2018neural,Wang_2018_ECCV} and implicit functions~\cite{chen2019learning,park2019deepsdf,mescheder2019occupancy}.
Those holistic representations, however, cannot capture part-level object structure, which lack fine-grained shape constrains and are sensitive to occlusion.

Primitive-based methods address those challenges by adopting a part-based representation. \cite{zou20173d} generates a primitive sequence directly from encoded image feature.
\cite{niu2018im2struct} explicitly encodes shape structure like adjacency and symmetry based on a binary tree of 3D boxes.
\cite{paschalidou2020learning} also learns to recover a binary tree of part-based shape decomposition.
Nevertheless, these methods generate object parts in a progressive manner and hence are prone to error propagation.
More recently, 
\cite{deng2020cvxnet} represents shapes via a learnable convex decomposition and \cite{chen2020bsp} proposes to generate shapes via binary shape partitioning, both utilizing a fully-connected network to predict all object parts at once. By contrast, we explicitly model part relations with a non-local graph neural network~\cite{wang2018non}, which enables us to incorporate a fine-grained shape regularity constraint, and to explore the guidance of part semantic relations in shape recovery with additional supervision.

Most existing methods adopt an encoder-decoder network architecture, which either directly encode the entire image into a feature representation~\cite{girdhar2016learning}, or resort to 2.5D intermediate representations such as silhouette, depth and normal map~\cite{wu2017marrnet,wu2018learning} for better input features, and then map it to 3D shape output.
However, such holistic features are susceptible to occlusion. In contrast, we propose to use attention mechanism to extract local features based on 2D bounding box predictions for object parts.


\subsection{Part-based 3D Shape Representation}
Representing objects with parts and discovering object structure is a long-standing research problem~\cite{mitra2014structure,wang2011symmetry,zheng2014recurring,han2020shapecaptioner}.
Recent approaches learn a part-based object representation by mapping a shape volume to a set of primitive boxes~\cite{tulsiani2017learning}, or superquadrics~\cite{paschalidou2019superquadrics}, or 3D Gaussians~\cite{genova2019learning,genova2020local}, but largely ignore object part relations and high-level shape regularity.
\cite{li2017grass} designs a recursive autoencoder in a binary tree to generate object parts, which explicitly captures part symmetry and adjacency. 
StructureNet~\cite{mo2019structurenet} enriches this tree structure into an n-ary graph and introduces part semantics in shape generation. 
\cite{tian2018learning} represents 3D volumes with programs~\cite{Sharma_2018_CVPR} and further combines semantic structure to capture shape regularity and stability of real-world objects. 
\cite{wu2020pq} proposes a sequence-to-sequence network for part-based object generation. However, those approaches typically generate part-based shapes in a progressive manner and hence lack global refinement on the 3D object shapes.
Moreover, they often use a simple encoder-decoder strategy without grounding part-based object structure into input image, and have difficulty in parsing complex realistic scenes.


\begin{figure*}[t!]
	\renewcommand\arraystretch{0.005}
	\begin{center}
		\scalebox{0.85}[0.85]{
			\begin{tabular}{c}
				\raisebox{-0.9\height}{\includegraphics[width=\textwidth]{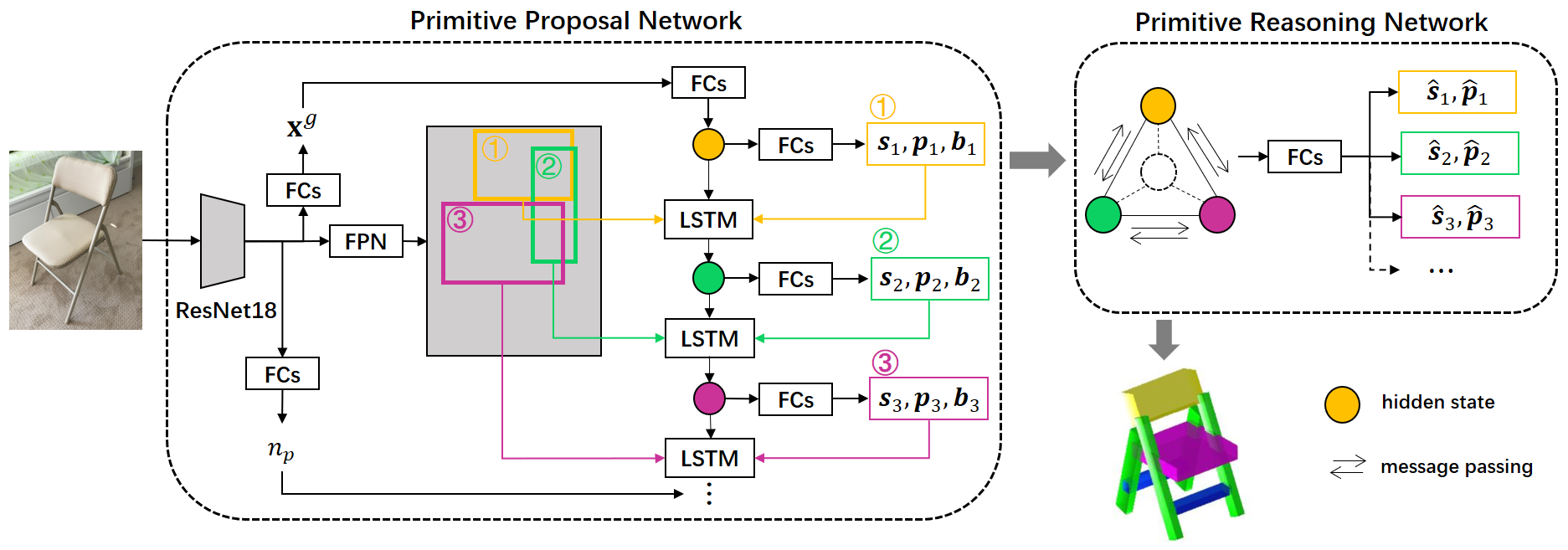}}
			\end{tabular}
		}
	\end{center}
	\caption{Model framework. Our model consists of two main modules: Primitive Proposal Network generates a sequence of proposals from the input image with LSTMs, and Primitive Reasoning Network is a fully-connected graph network built on the proposals to jointly reason the 3D shape configuration\protect\footnotemark.}
	\label{fig:model}
\end{figure*}

\section{Method}
\subsection{Problem Setting}\label{sec:setting}

Given an input RGB or depth image $I$ of an object $O$, our goal is to estimate its full 3D shape in a form of a set of 3D geometric primitives $\mathcal{P}=\{\mathbf{p}_1,...,\mathbf{p}_{N_O}\}$, where $N_O$ is the number of its primitives. In this work, we adopt a box representation for each primitive $\mathbf{p}_i=[l_x, l_y, l_z, t_x, t_y, t_z, \theta_x, \theta_y, \theta_z]\in\mathbb{R}^{9}$, where $[l_x, l_y, l_z]$ measures the length of three orthogonal edges of the box, $[t_x, t_y, t_z]$ represents the translation of the box center and $[\theta_x, \theta_y, \theta_z]$ denotes the rotated Euler angles along each axis. 

In order to incorporate object part semantics, we also consider the problem setting of 3D recovery with additional semantic part annotations for each object category $c$, in which a semantic part label $s_{i} \in \{1,...,M_c\}$ is assigned to each primitive $\mathbf{p}_i$. In the rest of the paper, we refer to this part-augmented scenario as the \textit{semantic-aware setting} while denoting the more general scenario without such annotations (i.e., $M_c=1$) as the \textit{semantic-agnostic setting}. 

\footnotetext{We only show one LSTM sequence in our proposal generation for clarity. Note that we also use global image feature $\mathbf{x}^g$ as input to LSTM at each step of proposal generation.}

\subsection{Model Overview}\label{sec:overview}
To tackle the problem of 3D primitive estimation from a monocular image, we adopt a hypothesize-and-refine strategy, in which we first generate a set of 3D primitive proposals from the image and then jointly infer their 3D parameters and semantic part labels. We instantiate this two-stage strategy by developing a modular graph neural network consisting of two main modules: a primitive proposal network and a primitive reasoning network. The front-end proposal network extracts the set of initial primitive proposals from the input image, while the next reasoning network captures global context of the entire object and performs the refinement of primitive features, which are used for final prediction on their geometry and semantic labels. Fig.~\ref{fig:model} shows an overview of our model architecture.   

In the remaining of this section, we will introduce our network design in details. We start from the semantic-aware setting and first introduce the primitive generation network in Sec.~\ref{sec:proposal}, followed by the primitive reasoning network in Sec~\ref{sec:graph}. We then discuss our stage-wise model learning in Sec.~\ref{sec:learn}. Finally, we generalize our method to the semantic-agnostic setting in Sec.~\ref{sec:agnostic}.

\subsection{Primitive Proposal Network}\label{sec:proposal}

Our first step is to generate a set of 3D box primitives from the input image. Generating each primitive independently from a 2D input is a challenging task and hence we consider a structural approach to exploit the 3D shape prior of objects. To this end, we introduce a CNN-RNN model that predicts a sequence of primitive proposals each time. Such a sequence proposal generation mechanism captures the dependency between neighboring primitives, and is capable of producing more stable 3D candidate boxes.   

Specifically, we propose a Primitive Proposal Network which consists of two submodules: a {Backbone ConvNet} and a {Recurrent Generator}. The {Backbone ConvNet} computes image features and estimates the number of primitives. Then a {Recurrent Generator} utilizes recurrent units to predicts primitive proposals, including their semantic labels and geometric parameters. Below we describe the details of those two submodules.

\subsubsection{Backbone ConvNet}
We use a ResNet18+FPN~\cite{lin2017feature} as the backbone network to compute a 2D feature map $\mathbf{\Gamma}$ for later local conv feature extraction. Then we attach two sets of fully-connected layers (FCs) to the ResNet18, separately, to obtain a global image representation $\mathbf{x}^g\in \mathbb{R}^{256}$ and to directly regress the number of object primitives $N_O$ for this instance. The output of the regressor is denoted as $n_p$ after rounding.

\subsubsection{Recurrent Generator}
We adopt a light-weight two-layer LSTM network to generate primitive proposals in a sequential manner. 
At each step $i$, the LSTM updates its hidden state $\mathbf{h}_i \in \mathbb{R}^{D_h}$ and output three properties of the current primitive, including its 3D parameters $\mathbf{p}_{i}$, semantic label distribution $\mathbf{s}_{i}\in \mathbb{R}^{M_c}$, and its 2D bounding box location $\mathbf{b}_{i}\in \mathbb{R}^4$ in image. 

The initial hidden state is computed from the global image feature $\mathbf{x}^g$ with an MLP as $\mathbf{h}_1 = f_{h}(\mathbf{x}^g)$, and the LSTM unit, denoted as $F_{\text{LSTM}}$, updates the hidden state as follows, 
\begin{align}
\mathbf{h}_{i+1} &= F_{\text{LSTM}}(\mathbf{h}_{i}, \mathbf{x}_{i}, \mathbf{s}_{i}, \mathbf{p}_{i}), \quad 1\leq i < n_p \label{eq:lstm}
\end{align}
where $\mathbf{x}_{i}$ is a set of input features including the global image feature $\mathbf{x}^g$, the bounding box location $\mathbf{b}_{i}$ and a local conv feature $\mathbf{v}_{i}$ extracted with ROI-Align~\cite{he2017mask} on the conv feature map $\mathbf{\Gamma}$.

Given the hidden state $\mathbf{h}_i$, the LSTM unit outputs three predictions on the current primitive with MLPs as follows, 
\begin{align}
\mathbf{s}_i &= f_{s}(\mathbf{h}_i), \;\;
\mathbf{b}_i = f_{b}(\mathbf{h}_i), \;\;
\mathbf{p}_i = f_{p}(\mathbf{h}_i, \hat{s}_i), \;\; \hat{s}_i=\arg\max(\mathbf{s}_i) \label{eq:readout}
\end{align}
where $f_{s}(\cdot)$ is a single FC followed by a Softmax function and $f_{b}(\cdot)$, $f_{p}(\cdot)$ are multiple FCs with ReLU as activation function. We use a different MLP for each semantic class $s_i$ to predict primitives.

As generating 3D primitives from 2D images is particularly challenging, we build two separate LSTM networks to improve the recall rate of our proposal set. Specifically, the networks generate two primitive sequences in opposite directions for each input image (see Sec.~\ref{sec:exp-detail}). We discard the ordering information in two sequences, and take their union as the output of the Primitive Proposal Network. We denote the final proposal set as $\mathcal{Q}=\{(\mathbf{p}_i,\mathbf{s}_i,\mathbf{h}_i)\}_{i=1}^{2n_p}$. 



\subsection{Primitive Reasoning Network}\label{sec:graph}
Given the generated proposal set $\mathcal{Q}$, our second module aims to refine the proposals jointly by taking into account the global context of the entire object shape and to generate the final prediction of 3D primitives. To this end, we propose a fully-connected graph neural network~\cite{wang2018non} to update the primitive representations, based on which we then predict the parameters of 3D primitives.    

Specifically, we first build a primitive graph $\mathcal{G}$ whose nodes are primitive proposals and edges are fully connected. We then associate the $ i $-th primitive to its corresponding node and denote its feature embedding as $\mathbf{z}_i\in \mathbb{R}^{D_z}$. The initial feature embedding is computed from the proposal features: 
\begin{align}
\mathbf{z}_i = [g_{h}(\mathbf{h}_i); g_{s}(\mathbf{s}_i); g_p(\mathbf{p}_i)], \quad 1\leq i \leq 2n_p \label{eq:node-feat}
\end{align}
where $g_{h}(\cdot)$,$g_{s}(\cdot)$,and $g_p(\cdot)$ are a FC layer followed by a ReLU function, respectively. $[;]$ denotes concatenation. 

Our primitive reasoning network uses a one-step message passing to update the primitive feature embedding for capturing the global context of each object instance. Concretely, given a graph node $\mathbf{z}_i$, we aggregate information from all the remaining primitives with an attention mechanism and update the node feature with a residual block:
\begin{align} 
\alpha_{ij} = \left(\mathbf{U}\mathbf{z}_i\right)^\intercal \left(\mathbf{V}\mathbf{z}_j\right),
\quad
\mathbf{y}_i &= \mathbf{z}_i + \frac{1}{2n_p-1}\sum_{\forall j\neq i} \alpha_{ij} \mathbf{W}\mathbf{z}_j
\end{align}
where $\mathbf{U}, \mathbf{V}, \mathbf{W} \in \mathbb{R}^{D_z\times D_z}$ are embedding matrices and  $\alpha_{ij}$ is the importance weight of each input message, and $\mathbf{y}_i$ is the updated primitive representation\footnote{We note that while it is straightforward to extend the message passing to multiple iterations, we found empirically that single iteration achieves the best performance.}.
 

\subsubsection{Model Prediction}
Given the updated primitive representations, we now refine the semantic label prediction and geometric parameters of all the proposals. To remove the false positive primitives, we augment the semantic label space with a background class $0$ so that the semantic class of the $i$-th primitive $s_i \in \{0,1,...,M_c\}$. The final prediction of the semantic label and the primitive parameters are estimated based on two FC networks as follows: 
\begin{align}
\hat{\mathbf{s}}_i &= f_{s}^o(\mathbf{y}_i), \quad
\hat{\mathbf{p}}_i = f_{p}^{o}(\mathbf{y}_i,\hat{s}_i), \quad
\hat{s}_i =\arg\max \hat{\mathbf{s}}_i \wedge \hat{s}_i>0 \label{eq:graph-readout}
\end{align}
where $f_{s}^{o}(\cdot)$ is a single FC layer followed by a softmax function and $f_{p}^{o}(\cdot)$ is a set of MLPs with ReLU as activation function, one for each part semantic class. 

During testing, we apply non-maximum suppression on the final primitive predictions. Specifically, we first match all primitives into pairs according to their L1 distance. Each time we match a pair with minimum distance in the remaining unpaired primitives. After the matching process, we take the primitive with higher class confidence score in each pair to be our final outputs. We use pairing for NMS as we have two LSTM models to generate proposals in two directions. Each LSTM typically generates non-overlapping primitives due to sequential dependency, and we only need to remove potential redundant proposals from two LSTMs.
\subsection{Model Learning}\label{sec:learn}
Based on our modular architecture, we adopt a stage-wise strategy in our model learning. Below we will introduce the training process of the Primitive Proposal Network and the Primitive Reasoning Network in turn. For notation clarity, we define our training loss on an input image of object $O$.  
\subsubsection{Training Primitive Proposal Network}
For our backbone network, we fix the ResNet18 with pre-trained parameters and finetune the parameters of the FPN module with the rest of the proposal network. The training of the regressor and the recurrent generator are decoupled due to the fixed base network. 

We first train the regressor with an $l_1$ loss on the predicted number of primitives, which is then applied to the next stage for training primitive reasoning module.
For the recurrent generator, we assume the primitives are generated with a predefined ordering and length $N_O$ as specified in the ground-truth (see Sec.~\ref{sec:exp-detail}). The overall loss for the proposal network consists of three terms: a semantic label classification loss, a primitive regression loss and a box regression loss, which are defined as follows:
\begin{align}\label{loss:s2}
L_{p} &= 
\sum_{i=1}^{2N_O} L_{ce}(\mathbf{s}_i, s_i^*) 
+ \lambda_{rp} \sum_{i=1}^{2N_O} L_{l_1}(\mathbf{p}_i, \mathbf{p}_i^*) 
+ \lambda_{rb} \sum_{i=1}^{2N_O} L_{sm}(\mathbf{b}_i,\mathbf{b}_i^*)
\end{align}
where $ s_i^* $, $ \mathbf{p}_i^* $ and $ \mathbf{b}_i^* $ are the ground-truth semantic label, primitive and box location, $ L_{ce} $,$ L_{l_1} $ and $ L_{sm} $ are Cross Entropy loss, L1 distance and Smooth-L1 loss, and $\lambda_{rp}$ and $\lambda_{rb}$ are the corresponding weights to primitive regression loss and box regression loss, respectively.

\subsubsection{Training Primitive Reasoning Network}
We train the primitive reasoning network with a multi-task loss similar to the object detection systems. 
Specifically, given $ 2n_p $ final predictions, we match them to $2N_O$ ground-truth primitives (doubled) in a greedy manner: Each time we first compute the $l_1$ distance between all remained prediction and ground-truth primitive pairs and choose the pair with minimum distance. Then we remove the matched pair from remained primitives and repeat matching, until no predictions or no ground-truth primitives are left. Finally, if any predictions remain unmatched, we assign them to background class. Empirically, we do not include rotation in this $l_1$ distance computation, as rotation predictions are relatively less reliable than primitive translation and edge length.

The total loss for the Primitive Reasoning Network has two terms, including a semantic classification loss and a primitive regression loss, which can be written as:
\begin{align}\label{loss:s3}
L_{r} = 
\sum_{i=1}^{2n_p} L_{ce}(\hat{\mathbf{s}}_i, \hat{s}_i^*) 
+ \lambda_{rp} \sum_{i=1}^{2n_p} L_{l_1}(\hat{\mathbf{p}}_i, \hat{\mathbf{p}}_i^*)\llbracket\hat{s}_i^*>0\rrbracket
\end{align}
Where $ \hat{s}_i^* $ and $ \hat{\mathbf{p}}_i^* $ are the ground-truth semantic label and primitive respectively, $\llbracket\cdot\rrbracket$ is the indicator function, and the weight $\lambda_{rp}$ of primitive regression loss is the same as in Eq.~\ref{loss:s2}.

\subsection{Semantic-agnostic Model}\label{sec:agnostic}
For the semantic-agnostic setting, it is straightforward to extend our model formulation by setting $ M_c = 1 $, which indicates the semantic part annotations are unavailable. 

In addition, we simplify our model slightly to speed up model inference and learning. Specifically, in the Primitive Proposal Network, as $ M_c = 1 $, we set the output function for the semantic label $f_s$ as a constant, i.e., $f_s(\cdot)\equiv 1$ and remove the semantic label classification loss in Eq.~\ref{loss:s2} in the training stage. Similarly, in the  Primitive Reasoning Network, we set the embedding function for the semantic label $g_s$ as a constant, i.e., $g_s(\cdot)\equiv 1$. We note that the rest of our model pipeline and training process are the same as the semantic-aware setting. 

\section{Experiment}
We train and evaluate our model on Pix3D dataset~\cite{sun2018pix3d} with real-world RGB images and ModelNet dataset~\cite{wu20153d} with synthetic depth images in both semantic-aware and semantic-agnostic setting. 
In addition, we directly apply the models trained on ModelNet to NYU Depth V2~\cite{silberman2012indoor} dataset to demonstrate the robustness of our method. 
On each dataset, we compare with the state-of-the-art methods 3D-PRNN~\cite{zou20173d}, Tulsiani et al.~\cite{tulsiani2017learning}, and PQ-Net~\cite{wu2020pq}.

Below we first introduce the datasets and evaluation protocol in Sec.~\ref{sec:exp-data} and implementation details in Sec.~\ref{sec:exp-detail}. Then we present results comparing to other methods on three benchmarks in Sec.~\ref{sec:exp-results}. 
To illustrate the effectiveness of different components of our model design, we conduct comprehensive ablation study on Pix3D \emph{chair} class in Sec.~\ref{sec:exp-ablation}. Finally, we demonstrate further analysis of our method in supplementary, regarding our proposal quality, robustness of our method, and our failure cases.

\subsection{Datasets and Metrics}\label{sec:exp-data}
\noindent\textbf{Pix3D}
Pix3D~\cite{sun2018pix3d} is a challenging benchmark consisting of pixel-level aligned 3d shapes and real-world images with various occlusion and background clutter. We conduct our experiments on its three major categories: \emph{chair}, \emph{table} and \emph{sofa}, which contain 3839, 1870 and 1947 images, as well as 221, 63 and 20 unique 3D models, respectively. 
We take all models in voxel of $32\times32\times32$ and manually label them into well-aligned primitives with corresponding semantic class. We assign class label to each part based on its functionality, such as cushion to sit on and legs to support. Given camera pose and primitive annotations, we obtain 2D part bounding boxes from reprojected 3D primitives. We divide all parts of \emph{chair} into six semantic classes, \emph{table} into four, and \emph{sofa} into four. The maximum number of primitives in an object for \emph{chair}, \emph{table}, and \emph{sofa} are 15, 8, and 8.
We randomly sample and fix 20\% of images to be our test set and use the rest as our training set. Note that our test and train splits have no shared 3D model annotations.

Our semi-automatic manual labeling of primitives consists of two steps: First, we utilize the same preprocessing toolbox as in the 3D-PRNN to generate a coarse primitive (oriented bounding box) configuration for each object volume in $32\times32\times32$ from original ground truth volume annotations. The toolbox is based on an energy minimization procedure that resembles Iterative Closest Point (ICP). In the second step, we then slightly adjust the parameters of those primitives manually by visually inspecting their alignment with the ground truth volume through a graphical user interface.

\noindent\textbf{ModelNet} 
We follow 3D-PRNN and apply our method on three categories of ModelNet~\cite{wu20153d}: \emph{chair}, \emph{table} and \emph{night stand}. For each category, we utilize 889, 392 and 200 models for training and 100, 100 and 82 models for testing, respectively. 
We adopt the same data synthesis strategy as in 3D-PRNN to get input depth maps of size $64 \times 64$. For each object, we sample five views on a unit sphere by rejection-sampling, bounded within 20 degrees of the equator, and compute depth by projection of meshes. 
The primitive annotations are from 3D-PRNN and we augment each primitive with a semantic class label. We divide all parts of \emph{chair} into six classes, \emph{table} into four, and \emph{night stand} into four. The maximum number of primitives in an object for \emph{chair}, \emph{table}, and \emph{night stand} are 16, 11, and 10.

\noindent\textbf{NYU Depth V2} 
NYU Depth V2~\cite{silberman2012indoor} is a very challenging indoor scene dataset consisting of real-world RGBD images.
We use annotations from~\cite{guo2013support}, with 30 CAD models of six furniture categories: \emph{chair}, \emph{table}, \emph{desk}, \emph{bed}, \emph{bookshelf} and \emph{sofa}, and extruded polygons for other objects.
As primitive annotations are not available, we directly apply the models trained on ModelNet to the test split of NYU Depth V2 and compare the results in voxel.


\begin{table}[t!]
	\caption{Performance on Pix3D~\cite{sun2018pix3d} dataset. Results for Tulsiani, 3D-PRNN, Ours Agn, and Ours Sem are for OBB estimation, while results for PQ-Net and Ours Sem* are for AABB. ${\rm IoU}_p$ indicates our comparison to the voxelized primitive ground truth, focusing on primitive quality instead of surface details.}
	\label{pix}
	\begin{center}
		\resizebox{0.42\textwidth}{!}{
			\begin{tabular}{lccccccccc}
				\toprule
				Methods & ${\rm HErr}$ & ${\rm TAcc}^{0.1}$ & ${\rm TAcc}^{0.2}$ & ${\rm TAcc}^{0.3}$ & ${\rm IoU}_p$ \\
				\midrule\midrule
				\textbf{\emph{Chair}} \\
				\midrule
				Tulsiani & - & - & - & - & 19.2 \\
				3D-PRNN & 0.242 & 5.8 & 26.1 & 50.6 & 25.9 \\
				Ours Agn & 0.207 & 6.7 & 31.2 & 55.6 & 30.0 \\
				Ours Sem & \textbf{0.187} & \textbf{10.5} & \textbf{42.5} & \textbf{64.6} & \textbf{38.1} \\
				\midrule
				PQ-Net & \emph{0.215} & \emph{6.8} & \emph{27.3} & \emph{48.2} & \emph{36.6} \\
				Ours Sem* & \emph{0.186} & \emph{10.9} & \emph{43.8} & \emph{66.9} & \emph{40.3}\\
				\midrule
				\textbf{\emph{Table}} \\
				\midrule
				Tulsiani & - & - & - & - & 8.4 \\
				3D-PRNN & 0.295 & 3.8 & 29.1 & 56.0 & 12.0 \\
				Ours Agn & 0.245 & 8.2 & 41.9 & 64.5 & 19.6 \\
				Ours Sem & \textbf{0.233} & \textbf{12.8} & \textbf{43.0} & \textbf{70.5} & \textbf{20.5} \\
				\midrule
				PQ-Net & \emph{0.247} & \emph{8.3} & \emph{38.1} & \emph{62.2} & \emph{19.0} \\
				Ours Sem* & \emph{0.230} & \emph{12.8} & \emph{43.8} & \emph{72.7} & \emph{20.6}\\
				\midrule
				\textbf{\emph{Sofa}} \\
				\midrule
				Tulsiani & - & - & - & - & 55.4 \\
				3D-PRNN & 0.073 & 73.5 & 91.1 & 94.4 & 74.1 \\
				Ours Agn & 0.061 & 76.7 & 95.9 & 97.3 & 76.6 \\
				Ours Sem & \textbf{0.055} & \textbf{78.7} & \textbf{98.8} & \textbf{99.3} & \textbf{79.6} \\
				\midrule
				PQ-Net & \emph{0.065} & \emph{67.7} & \emph{96.6} & \emph{98.2} & \emph{73.2} \\
				Ours Sem* & \emph{0.054} & \emph{78.7} & \emph{98.8} & \emph{99.5} & \emph{79.6} \\
				\bottomrule
			\end{tabular}
		}
	\end{center}
	
\end{table}

\noindent\textbf{Evaluation Protocol}
For our semantic-aware setting, we train a class-specific model with semantic supervision for each object category. While for our semantic-agnostic setting, we only train one class-agnostic model with no semantics on all object categories. 

We compare with the state-of-the-art methods in two different settings. 
Firstly, we compare our method with 3D-PRNN~\cite{zou20173d} and Tulsiani et al.~\cite{tulsiani2017learning}, for estimating Oriented Bounding Box (OBB) primitives from single images. Note that Tulsiani et al.~\cite{tulsiani2017learning} uses mesh as supervision. Secondly, we compare our method with PQ-Net~\cite{wu2020pq} in recovering Axis-Aligned Bounding Box (AABB) primitives, which outperforms other methods, such as StructureNet~\cite{mo2019structurenet}.

Specifically, we first compare our method in the semantic-agnostic setting (\textbf{Ours Agn}) with 3D-PRNN and Tulsiani, in recovering OBB. 
To show the benefits of semantic relations in both primitive reasoning and class-specific model setting, we also compare our semantic-aware model (\textbf{Ours Sem}) to the semantic-agnostic version. 
Moreover, we compare our semantic-aware model (\textbf{Ours Sem*}) with PQ-Net, which is also trained in class-specific setting, in recovering AABB. 

We evaluate our method using three metrics: \emph{Hausdorff error} (${\rm HErr}$), \emph{thresholded accuracy} (${\rm TAcc}^\delta$ [\%]) and \emph{intersection over union} (${\rm IoU}$ [\%]). To evaluate the quality of our estimated primitives, we adopt ${\rm HErr}$ and ${\rm TAcc}^\delta$ ($\delta$ is the threshold) from Im2Struct~\cite{niu2018im2struct}, which are based on Hausdorff distance of primitive pairs. Since the primitives are well-aligned in voxel space, we also voxelize our primitive predictions and compare them with the ground truth voxels using ${\rm IoU}$, to evaluate the quality of each object as a whole.

Given the test set of $T$ samples, we compute Hausdorff error 
${\rm HErr} = \frac{1}{2T} \sum_{i}^{T} (D(\mathcal{S}_i, \mathcal{S}_i^*) + D(\mathcal{S}_i^*, \mathcal{S}_i))$, 
where 
$\mathcal{S}_i$ is a predicted shape consisting of a set of primitives and $\mathcal{S}_i^*$ is its corresponding ground truth.
For each primitive, we represent it as a set $\mathcal{V}$, consisting of its eight corners.
$D(\mathcal{S}_1, \mathcal{S}_2) = \frac{1}{n} \mathop {\sum }\limits_{\mathcal{V}_1^j \in \mathcal{S}_1} \mathop {\min }\limits_{\mathcal{V}_2^k \in \mathcal{S}_2} H(\mathcal{V}_1^j, \mathcal{V}_2^k)$
is the averaged minimum Hausdorff distance from primitives in $\mathcal{S}_1$ to those in $\mathcal{S}_2$, where $n$ is the number of primitives in $\mathcal{S}_1$.
$H(\mathcal{V}_1, \mathcal{V}_2) = \mathop {\max }\limits_{\mathbf{q}_1 \in \mathcal{V}_1} \mathop {\min }\limits_{\mathbf{q}_2 \in \mathcal{V}_2} ||\mathbf{q}_1 - \mathbf{q}_2||$ is the Hausdorff distance between two vertex set $\mathcal{V}_1$ and $\mathcal{V}_2$. 

Thresholded accuracy ${\rm TAcc}^\delta$ is the percentage of predicted primitives that $H(\mathcal{V}_i, \mathcal{V}_i^*) / L(\mathcal{V}_i^*) < \delta$, 
where $\mathcal{V}_i$ is a primitive in recovered shape, 
$\mathcal{V}_i^*$ is its nearest primitive in ground truth shape in Hausdorff distance, 
$L(\mathcal{V}_i^*)$ is the length of the diagonal of primitive $\mathcal{V}_i^*$, and $\delta$ is the threshold.

\subsection{Implementation Details} \label{sec:exp-detail}
We normalize all primitive parameters to have zero mean and standard deviation. Following 3D-PRNN, we assume all objects have bilateral symmetrical primitive pairs. For such symmetrical pairs, we only model the left half in our network and generate the right half by mirroring w.r.t. to the symmetry plane, to improve learning efficiency.
For each object, we pre-sort all primitives based on the height of their centers, and use the proposal network to generate the primitive sequence in both bottom-up and top-down orderings separately.
The dimension of LSTM hidden state $D_h$ and graph node $D_z$ is 800 and 1024, respectively. The weights to primitive regression loss and box regression loss $\lambda_{rp}$ and $\lambda_{rb}$ are both 10. 

We re-implement 3D-PRNN and apply it to Pix3D and ModelNet datasets following their released codebase. 
For experiments of 3D-PRNN on Pix3D~\cite{sun2018pix3d}, we replace their original depth encoder with the same ResNet18 backbone as in our method, to encode RGB image features. 
To obtain AABB ground truth and predictions for \textbf{Ours Sem*}, we directly transform OBB ground truth and predictions from \textbf{Ours Sem}, by simply taking the tightest axis-aligned bounding box of each OBB. 
We train PQ-Net on AABB supervisions with their released code. 

Please find more implementation details in supplementary.\footnote{Our code is available at \href{https://github.com/hailieqh/3D-Object-Primitive-Graph}{https://github.com/hailieqh/3D-Object-Primitive-Graph}.}

\subsection{Results}\label{sec:exp-results}
We present comparison results with previous methods in two different settings: recovering Oriented Bounding Box (OBB) and Axis-Aligned Bounding Box (AABB).
Below we introduce the detailed results on three benchmarks.

\subsubsection{Pix3D} As shown in Table~\ref{pix}, for OBB estimation, \textbf{Ours Agn} outperforms 3D-PRNN in all metrics on all object categories. Our method achieves larger improvement on \emph{chair} and \emph{table}, whose shapes have more flexibility and are harder to recover compared to \emph{sofa}. This demonstrates that our model can capture stronger shape regularity and improve recovered shape quality by utilizing local features and joint primitive reasoning. 
Additionally, \textbf{Ours Sem} achieves better performance on all categories than \textbf{Ours Agn},
especially in challenging category \emph{chair} with a large margin (8.1\% higher in ${\rm IoU}_p$ and 0.020 lower in ${\rm HErr}$).
This is because \emph{chair} in Pix3D has much more geometric and semantic part variations than other categories, which shows that our model can utilize rich semantic relations to help handle large geometric variations. 
For AABB estimation, \textbf{Ours Sem*} also achieves better results than PQ-Net on all three categories.

\subsubsection{ModelNet} 
\begin{figure*}[t!]
	\huge
	\renewcommand\arraystretch{0.005}
	\begin{center}
		\resizebox{\textwidth}{!}{
			\begin{tabular}{c c c c c c c c c c c c c c}
				\raisebox{-0.9\height}{\includegraphics[width=\textwidth]{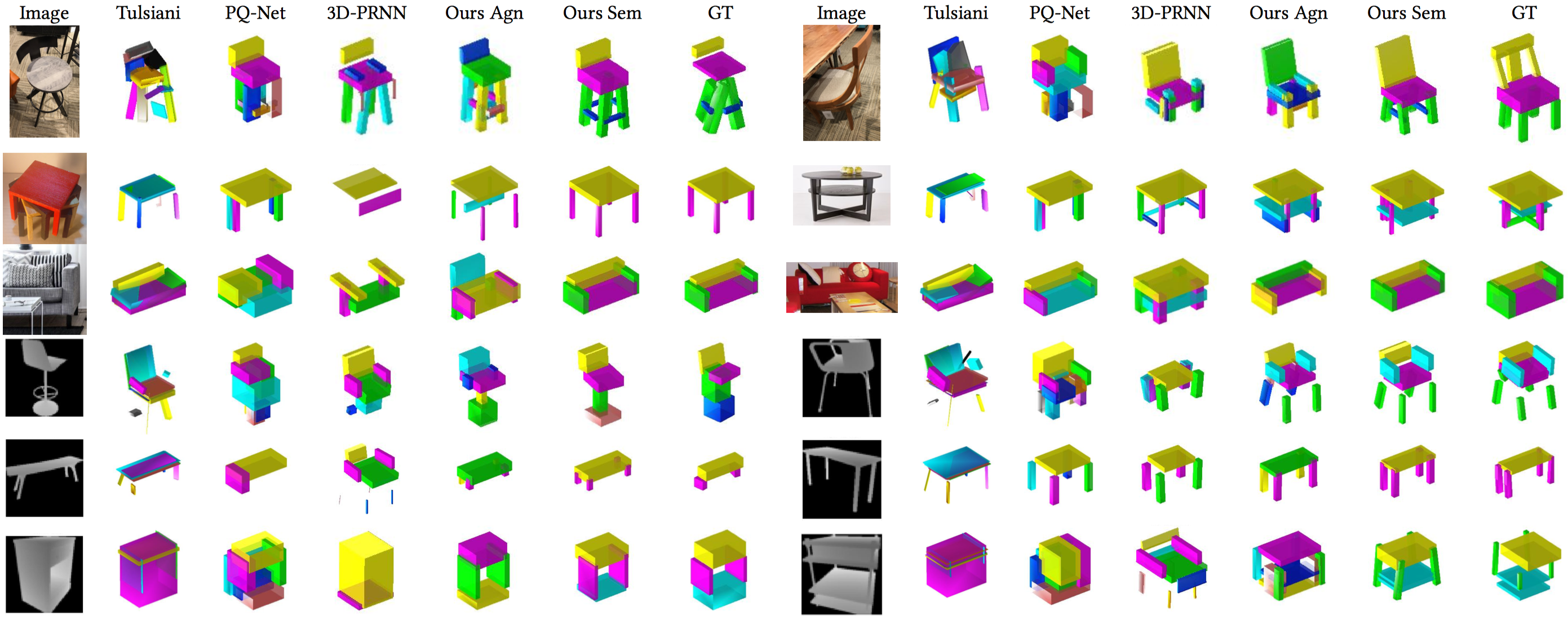}}
				
			\end{tabular}
		}
	\end{center}
	\caption{Qualitative results on Pix3D~\cite{sun2018pix3d} and ModelNet~\cite{wu20153d}. The color codings of Tulsiani, PQ-Net, 3D-PRNN, and Ours Agn are based on the order of drawing primitives and have no semantic meaning. The color codings of Ours Sem and GT are based on the semantics of primitives. Row 1-3 (Pix3D RGB): \emph{Chair}, \emph{Table}, \emph{Sofa}; Row 4-6 (ModelNet Depth): \emph{Chair}, \emph{Table}, \emph{Night stand}.}
	\label{fig:modelnet}
\end{figure*}

As shown in Table~\ref{modelnet}, for OBB estimation, 
\textbf{Ours Agn} achieves better performance than 3D-PRNN on all categories in ModelNet. The performance gap is especially large on \emph{night stand}, which has the least data for training. This shows that our model encodes stronger shape regularity prior with limited training data and thus can handle the problem of data imbalance more robustly. For \emph{table} class, our ${\rm IoU}_v$ is close to 3D-PRNN, but we achieves higher ${\rm IoU}_p$ by 4.4\% and lower ${\rm HErr}$ by 0.025, which is more related to the quality of our primitive prediction instead of detailed geometric surface. 
Moreover, \textbf{Ours Sem} also outperforms \textbf{Ours Agn} on all categories, but the gaps are not as large as those on Pix3D. 
This is because the primitives in ModelNet have larger distribution variation than Pix3D, hence they cannot be easily captured in our designed semantic space with a few labels.
For AABB estimation, \textbf{Ours Sem*} significantly outperforms PQ-Net in Hausdorff metrics ${\rm HErr}$ and ${\rm TAcc}^\delta$. 
The performance gaps are smaller in ${\rm IoU}_{p/v}$ for \emph{night stand}, which is mainly because erroneous primitives have less impact on voxel, especially for object categories with large volume and regular shapes.
The results on ModelNet verifies the capability of our method in modeling complex primitive dependency, as objects in ModelNet have much larger variations of primitive combinations than objects in Pix3D. (See also Fig.~\ref{fig:modelnet} for visual examples.)

\subsubsection{NYU Depth V2} We train our model on ModelNet synthetic data and directly applied to NYU Depth V2 real-world depth images. 
Despite the large domain gap between training and testing, we outperform 3D-PRNN on all categories for OBB estimation, as shown in Table~\ref{nyu}. We notice that the performance gaps on \emph{chair} and \emph{table} are relatively small. 
This is mainly because the sequential model cannot provide high quality proposals when heavy occlusion occurs, which further leads to suboptimal performance of our subsequent primitive reasoning module.
Note that our method still achieves better performance on \emph{night stand} with a considerable margin, which shows the robustness of our model in handling categories with limited training data.
Furthermore, \textbf{Ours Sem} outperforms \textbf{Ours Agn} with a considerable margin, demonstrating the benefits of semantic-aware models for extremely challenging cases in real-world scenarios. 
For AABB estimation, \textbf{Ours Sem*} also outperforms PQ-Net on \emph{chair} and \emph{table}. 


\begin{table}[t!]
	\large
	\caption{Performance on ModelNet~\cite{wu20153d}. In addition to ${\rm IoU}_p$, we also evaluate primitives in ${\rm IoU}_v$, where $v$ denotes we compare our voxelized primitive predictions to original ground truth voxel. We show both the originally reported results and our re-implementation of 3D-PRNN before and after '/'.}
	\label{modelnet}
	\begin{center}
		\resizebox{0.48\textwidth}{!}{
			\begin{tabular}{lccccccccc}
				\toprule
				Methods & ${\rm HErr}$ & ${\rm TAcc}^{0.1}$ & ${\rm TAcc}^{0.2}$ & ${\rm TAcc}^{0.3}$ & ${\rm IoU}_p$ & ${\rm IoU}_v$ \\
				\midrule\midrule
				\textbf{\emph{Chair}} \\
				\midrule
				Tulsiani & - & - & - & - & 30.4 & 26.8 \\
				\cite{wu20153d} & - & - & - & - & - & 25.3 \\
				3D-PRNN & -/0.176 & -/5.0 & -/26.2 & -/55.1 & -/40.1 & 23.8/28.6 \\
				Ours Agn & 0.160 & 6.6 & 29.5 & 54.6 & 44.4 & 32.2 \\
				Ours Sem & \textbf{0.157} & \textbf{8.7} & \textbf{36.2} & \textbf{59.9} & \textbf{46.5} & \textbf{33.6} \\
				\midrule
				PQ-Net & \emph{0.210} & \emph{1.4} & \emph{11.7} & \emph{35.4} & \emph{34.7} & \emph{22.0} \\
				Ours Sem* & \emph{0.156} & \emph{11.0} & \emph{39.2} & \emph{64.2} & \emph{47.9} & \emph{31.2} \\
				\midrule
				\textbf{\emph{Table}} \\
				\midrule
				Tulsiani & - & - & - & - & 30.3 & 22.4 \\
				\cite{wu20153d} & - & - & - & - & - & 25.0 \\
				3D-PRNN & -/0.190 & -/12.4 & -/38.0 & -/66.6 & -/34.0 & 26.3/28.9 \\
				Ours Agn & 0.165 & 16.8 & 37.7 & 59.6 & 38.4 & 29.7 \\
				Ours Sem & \textbf{0.160} & \textbf{20.1} & \textbf{42.9} & \textbf{64.7} & \textbf{41.9} & \textbf{33.3} \\
				\midrule
				PQ-Net & \emph{0.227} & \emph{8.6} & \emph{20.7} & \emph{37.5} & \emph{27.0} & \emph{19.7} \\
				Ours Sem* & \emph{0.160} & \emph{20.7} & \emph{45.2} & \emph{68.2} & \emph{41.5} & \emph{33.0} \\
				\midrule
				\textbf{\emph{Night stand}} \\
				\midrule
				Tulsiani & - & - & - & - & 42.5 & \textbf{42.9} \\
				\cite{wu20153d} & - & - & - & - & - & 29.5 \\
				3D-PRNN & -/0.279 & -/4.2 & -/24.5 & -/48.7 & -/32.8 & 26.6/31.9 \\
				Ours Agn & 0.244 & 7.6 & 33.2 & 56.4 & 41.2 & 39.9 \\
				Ours Sem & \textbf{0.235} & \textbf{13.1} & \textbf{42.3} & \textbf{63.9} & \textbf{43.1} & 41.5 \\
				\midrule
				PQ-Net & \emph{0.285} & \emph{1.9} & \emph{16.8} & \emph{41.0} & \emph{39.6} & \emph{38.5} \\
				Ours Sem* & \emph{0.234} & \emph{13.1} & \emph{43.0} & \emph{64.8} & \emph{43.4} & \emph{41.9} \\
				\bottomrule
			\end{tabular}
		}
	\end{center}
	
\end{table}

\begin{table}[t!]
	\large
	\caption{Performance on NYU Depth V2~\cite{silberman2012indoor}. We evaluate in ${\rm IoU}_v$ as there are no primitive annotations and we compare our prediction to ground truth voxel.}
	\label{nyu}
	\begin{center}
		\resizebox{0.48\textwidth}{!}{
			\begin{tabular}{lcccccccccccccccccc}
				\toprule
				Category & 3D-PRNN & Ours Agn & Ours Sem & PQ-Net & Ours Sem* \\
				\midrule
				\midrule
				Chair & 13.8 & 14.6 & \textbf{17.2} & \emph{13.1} & \emph{16.9} \\
				Table & 5.2 & 6.0 & \textbf{8.1} & \emph{6.8} & \emph{8.2} \\
				Night stand & 8.6 & 18.7 & \textbf{28.0} & \emph{31.7} & \emph{28.3} \\
				\bottomrule
			\end{tabular}
		}
	\end{center}
	
\end{table}

\begin{table*}[t!]
	\caption{Ablation study on Pix3D~\cite{sun2018pix3d} chair. 1 or 2 under LSTM denotes the number of LSTM sequence models used to generate primitive proposals. We conduct the experiments in a drop-one-out manner. ${\rm TAcc}^{\delta}$ and ${\rm IoU}_p$ are in \%.}
	\label{ablation}
	\begin{center}
		\resizebox{0.785\textwidth}{!}{
			\begin{tabular}{ccccccccccccccccc}
				\toprule
				& \multicolumn{4}{c}{Module} & \multicolumn{5}{c}{Metric} \\

				Methods & LSTM & Sem & BConv & Graph & ${\rm HErr} \downarrow$ & ${\rm TAcc}^{0.1} \uparrow$ & ${\rm TAcc}^{0.2} \uparrow$ & ${\rm TAcc}^{0.3} \uparrow$ & ${\rm IoU}_p \uparrow$ \\ 
				\midrule
				\midrule

				Ours Sem
				&2&\checkmark&\checkmark&Dense
				&\textbf{0.187}&\textbf{10.5}&\textbf{42.5}
				&\textbf{64.6}&\textbf{38.1} \\

				\midrule

				
				\multirow{2}{*}{} 
				&2&\checkmark&\checkmark&Star-like
				&0.195&10.2&40.5&62.6&36.5\\
				&2&\checkmark&\checkmark&K-nearest
				&0.197&10.5&39.6&61.7&35.8\\
				&2&\checkmark&\checkmark&-
				&0.199&9.4&35.5&56.0&36.6\\
				&2&\checkmark&-&Dense
				&0.195&10.5&39.2&60.9&35.8\\
				&2&-&\checkmark&Dense
				&0.192&9.1&38.6&60.9&36.7\\
				&1&\checkmark&\checkmark&Dense
				&0.190&9.7&38.0&63.1&37.3\\
				&Faster R-CNN&\checkmark&\checkmark&Dense
				&0.203&9.2&35.5&58.7&33.2\\
				\midrule
				Baseline
				&2&-&-&-
				&0.211&6.4&28.7&48.4&34.0\\
				\bottomrule
				
			\end{tabular}
		}
	\end{center}
	
\end{table*}

\subsubsection{Qualitative Results}
We also visually compare the results of Tulsiani, PQ-Net, 3D-PRNN, \textbf{Ours Agn} and \textbf{Ours Sem}, as shown in Fig.~\ref{fig:modelnet}. 
For Pix3D real-world images, our models can more robustly handle occlusion, truncation, unusual viewpoints and challenging novel instances, compared to Tulsiani, PQ-Net and 3D-PRNN. 
For ModelNet synthetic data, our recovered shapes are also better in terms of both detailed shape consistency with images and global shape regularity. 
Note that on categories with limited training data such as \emph{night stand}, our models are able to correctly recognize object categories and to recover shapes matched to input images, while 3D-PRNN tends to generate an instance from the dominant category (\emph{chair}). 
This shows that our primitive reasoning can learn to capture stronger shape structure prior during training and thus can handle the problem of training data imbalance more robustly. For more visualization, please see supplementary.

\subsection{Ablation Study}\label{sec:exp-ablation}

In this section, we conduct several experiments to show the efficacy of the components of our full model \textbf{Ours Sem} on Pix3D \emph{chair} class, which has large instance variations in complex real-world environment, as shown in Table~\ref{ablation}.

\textbf{LSTM}:
Our bidirectional sequential proposal network captures strong primitive dependency within an object and generates better primitive proposals. 
With only one \textbf{LSTM} unit, our model performance drops by 0.003 higher in ${\rm HErr}$ and 0.8\% lower in ${\rm IoU}_p$. 
Moreover, we also explored using a Faster R-CNN backbone instead of \textbf{LSTM}. 
Specifically, we first train a Faster R-CNN on 2D bounding boxes to generate 2D proposals and then extract local conv features from 2D proposals to regress corresponding 3D primitive proposal parameters. 
In this case, model performance drops significantly by 0.016 higher in ${\rm HErr}$ and 4.9\% lower in ${\rm IoU}_p$.

\textbf{Sem}:
Semantic class supervision and semantic-aware reasoning helps reduce primitive output search space and provide more guidance for object structure recovery. 
Without \textbf{Sem}, we observe a performance drop, which is 0.005 higher in ${\rm HErr}$ and 1.4\% lower in ${\rm IoU}_p$ than \textbf{Ours Sem}.

\textbf{BConv}:
Local conv features extracted from box regions in an image can help handle occlusion and improve the robustness of our model. 
Without \textbf{BConv}, the performance of our model drops by 0.008 higher in ${\rm HErr}$ and 2.3\% lower in ${\rm IoU}_p$.

\textbf{Graph}:
Our graph network helps capture the global shape context and mitigate errors in proposal generation. 
Without \textbf{Graph}, we fully train the primitive proposal network and apply NMS on final predictions to remove duplicate primitives. In this case, model performance deteriorates by 0.012 higher in ${\rm HErr}$ and 1.5\% lower in ${\rm IoU}_p$. Note that the performance drops more significantly in ${\rm HErr}$ and ${\rm TAcc}^\delta$ than in ${\rm IoU}_p$. This is mainly because the fine-level primitive quality is much lower without joint reasoning and global refinement, while the coarse-level shapes after voxelization are less affected by erratic primitives. 
To investigate the impact of graph structure, we also conduct experiments on two variants of graph design: (1) a distance graph in which each node is connected to K-nearest (K=3) nodes based on primitive center distance; (2) a star-like graph in which every node is connected to the central cushion nodes. 
Our densely-connected graph outperforms those two sparse graphs in all the metrics.

\section{Conclusion}
In this paper, we have developed a primitive-based graph neural network for 3D object estimation from single images. Our method first uses a proposal network to lift the image features into a pool of primitive proposals, and then builds a fully-connected graph network on these proposals to refine primitive features. 
Besides, we optionally take into account semantic part cues to provide better guidance for object shape.
As a result, our method is able to conduct joint reasoning on the primitives for coherent 3D recovery.
Moreover, we adopt a stage-wise strategy in our model learning, in which it first learns a recurrent network for primitive generation and then trains a graph network to propagate messages between the primitives. Evaluations on three benchmarks show that our approach consistently outperforms prior approaches with a sizable margin. 
Particularly, our method can robustly handle (self-)occlusion and challenging viewpoints in real-world scenarios.


\bibliographystyle{ACM-Reference-Format}
\balance
\bibliography{egbib}

\clearpage
\appendix

\section{Implementation Details}
We choose ResNet18 as backbone for its strong performance and extensive usage in the prior work~\cite{mescheder2019occupancy,paschalidou2020learning}. Empirically, we also found that deeper ResNets provide little improvement.
To handle different inputs, we adopt the same network architecture, ResNet18+FPN, as our backbone, which takes three-channel images as input. As a result, we process the RGB and depth inputs in slightly different manners. For experiments on Pix3D, we directly send the input RGB images into the network, while for the ModelNet and NYUv2 dataset, we duplicate each depth image three times to build a 3-channel input for the network. Our ResNet18 module is pretrained on ImageNet~\cite{deng2009imagenet}.

We first train the proposal network for 20 epochs and then freeze the parameters during the training of primitive reasoning, to prevent it from overfitting on training data and providing no training signals for the primitive reasoning network. We train our model with batch size 16 for a maximum of 400 epochs and with an Adam~\cite{kingma2014adam} optimizer, whose learning rate is 1e-4 and betas are (0.95, 0.999). We explore batch size from 4 to 64, epochs from 200 to 500, and learning rate from 1e-2 to 1e-5. Final decisions on hyper-parameters are based on the performance on validation split. Then we fix all hyper-parameters to train on training and validation splits together to obtain our final models.

The implementation of our method is based on PyTorch 1.0.0. We run experiments on single TITAN Xp GPU cards, using less than 2GB GPU memory, in Ubuntu 16.04.4. It takes about four days to train a class-agnostic model on all three categories of Pix3D or ModelNet. The seeds to randomness are all fixed to 0.

\section{More Qualitative Results}
\begin{figure*}[t!]
	\huge
	\renewcommand\arraystretch{0.005}
	\begin{center}
		\resizebox{\textwidth}{!}{
			\begin{tabular}{c c c c c c c c c c c c c c c c c c}
				\raisebox{-0.9\height}{\includegraphics[width=\textwidth]{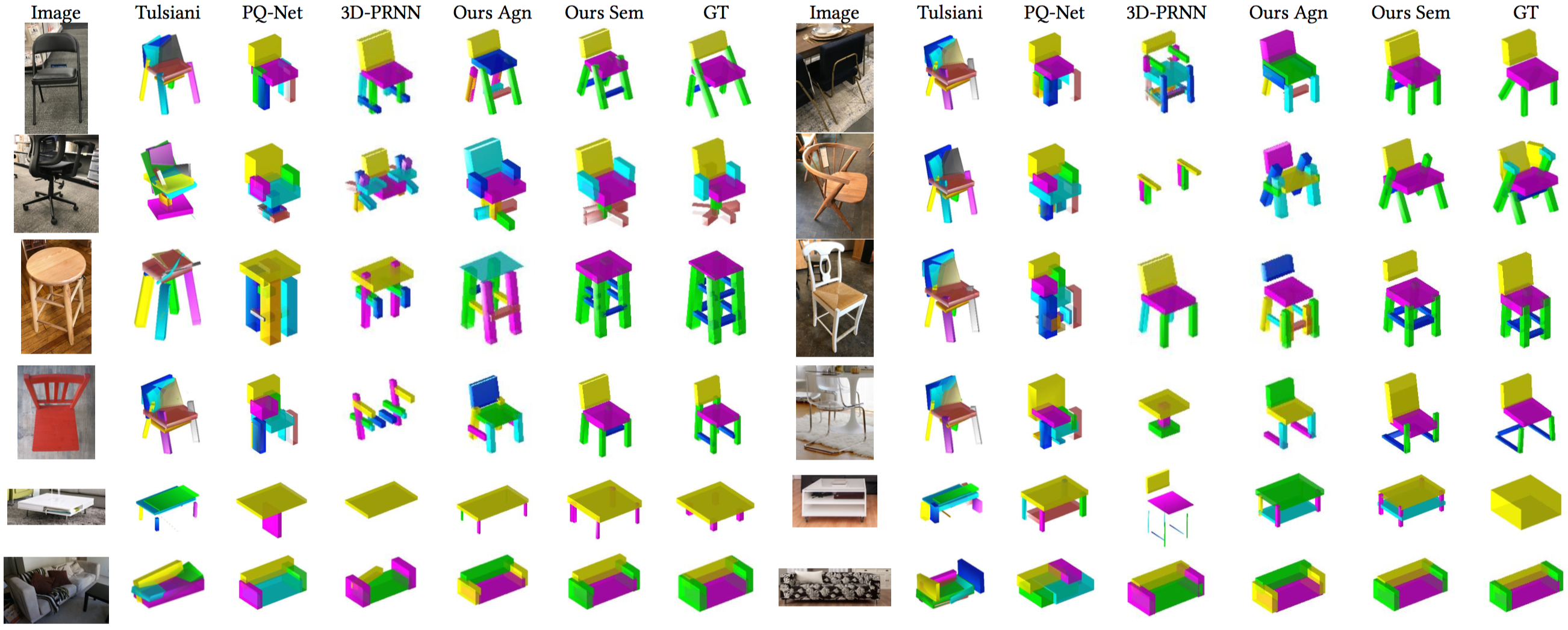}}
			\end{tabular}
		}
	\end{center}
	\caption{Qualitative results on Pix3D~\cite{sun2018pix3d}. Note that the color codings of Tulsiani, PQ-Net, 3D-PRNN, and Ours Agn are based on the order of drawing primitives and have no semantic meaning. In contrast, the color codings of Ours Sem and GT are based on the semantics of primitives.}
	\label{fig:pix}
\end{figure*}

\begin{figure*}[t!]
	\huge
	\renewcommand\arraystretch{0.005}
	\begin{center}
		\resizebox{\textwidth}{!}{
			\begin{tabular}{c c c c c c c c c c c c c c}
				\raisebox{-0.9\height}{\includegraphics[width=\textwidth]{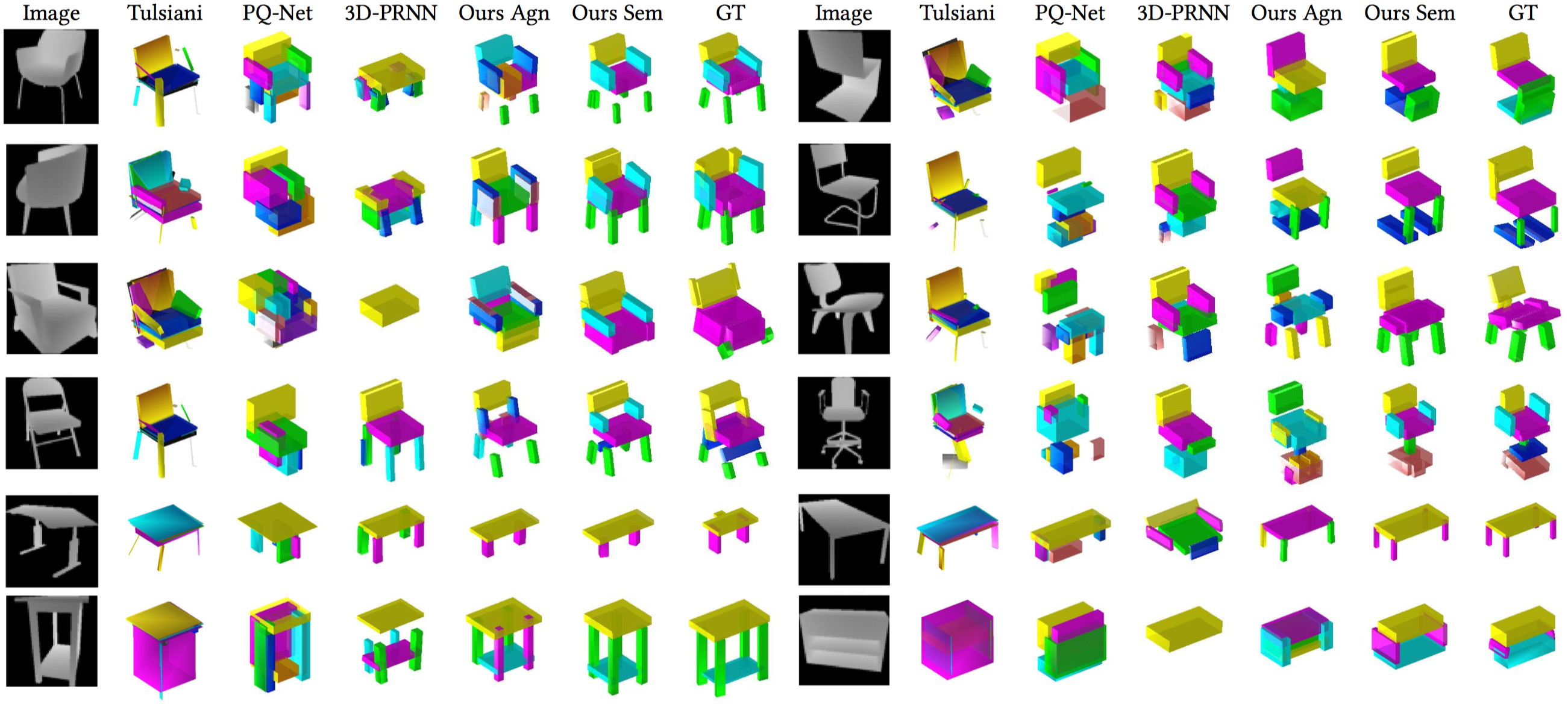}}

			\end{tabular}
		}
	\end{center}
	\caption{Qualitative results on ModelNet~\cite{wu20153d}. Note that the color codings of Tulsiani, PQ-Net, 3D-PRNN, and Ours Agn are based on the order of drawing primitives and have no semantic meaning. In contrast, the color codings of Ours Sem and GT are based on the semantics of primitives.}
	\label{fig:modelnet_suppl}
\end{figure*}

We also visually compare the results of Tulsiani, PQ-Net, 3D-PRNN, \textbf{Ours Agn} and \textbf{Ours Sem}, as shown in Fig.~\ref{fig:pix} for Pix3D and Fig.~\ref{fig:modelnet_suppl} for ModelNet. 
For Pix3D real-world images, our models can more robustly handle occlusion, truncation, unusual viewpoints and challenging novel instances, compared to Tulsiani, PQ-Net and 3D-PRNN. 
For ModelNet synthetic data, our recovered shapes are also better than Tulsiani, PQ-Net and 3D-PRNN, in terms of both detailed shape consistency with images and global shape regularity. 

\begin{table}[t!]
	\large
	\caption{Performance on different subsets of Pix3D~\cite{sun2018pix3d} chair. Set A, B, C, and D are truncated, occluded, slightly occluded, and complete (neither truncated nor occluded), respectively.} 
	\label{pix_splits}
	\begin{center}
		\resizebox{0.48\textwidth}{!}{
			\begin{tabular}{lccccccccc}
				\toprule
				& \multicolumn{4}{c}{${\rm HErr}\downarrow$} & \multicolumn{4}{c}{${\rm IoU}_p\uparrow$} \\
				Methods & Set A & Set B & Set C & Set D & Set A & Set B & Set C & Set D \\
				\midrule\midrule
				3D-PRNN & 0.261 & 0.249 & 0.311 & 0.233 & 21.0 & 23.3 & 17.0 & 27.8 \\
				Ours Sem & \textbf{0.176} & \textbf{0.186} & \textbf{0.249} & \textbf{0.184} & \textbf{40.8} & \textbf{38.1} & \textbf{30.7} & \textbf{38.2} \\
				\midrule
				PQ-Net & \emph{0.221} & \emph{0.220} & \emph{0.265} & \emph{0.210} & \emph{35.4} & \emph{36.5} & \emph{31.7} & \emph{37.3} \\
				Ours Sem* & \emph{0.175} & \emph{0.185} & \emph{0.247} & \emph{0.183} & \emph{42.1} & \emph{40.1} & \emph{32.6} & \emph{40.6} \\
				\bottomrule
			\end{tabular}
		}
	\end{center}
	
\end{table}

\begin{figure}[t!]
	\Large
	\renewcommand\arraystretch{0.005}
	\begin{center}
		\resizebox{0.48\textwidth}{!}{
			\begin{tabular}{c c c c c c c c c c c c}
				Image & Ours Sem & GT & Image & Ours Sem & GT\\
				\raisebox{-0.9\height}{\includegraphics[width=0.11\textwidth]{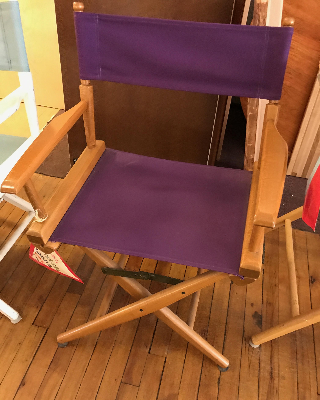}}
				&\raisebox{-0.9\height}{\includegraphics[width=0.12\textwidth]{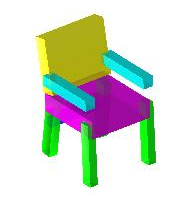}}
				&\raisebox{-0.9\height}{\includegraphics[width=0.12\textwidth]{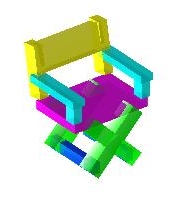}}
				
				&\raisebox{-1.1\height}{\includegraphics[width=0.12\textwidth]{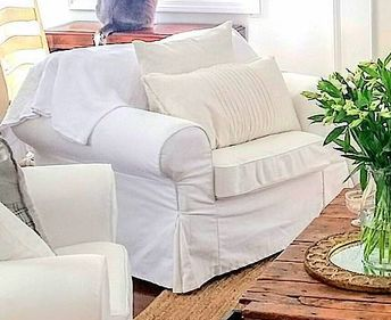}}
				&\raisebox{-0.9\height}{\includegraphics[width=0.12\textwidth]{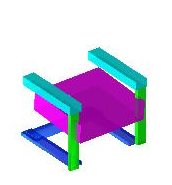}}
				&\raisebox{-0.9\height}{\includegraphics[width=0.12\textwidth]{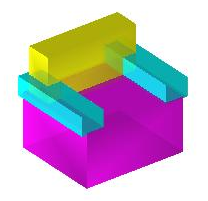}}\\
			\end{tabular}
		}
	\end{center}
	\caption{Two failure cases of our method on Pix3D chair. The chair on the left has bilaterally crossed legs which are unique in the dataset. The chair on the right has few texture and is occluded by objects with very similar appearance.}
	\label{fig:failure_case}
\end{figure}

\begin{table*}[t!]
	\caption{Ablation of proposal networks on Pix3D~\cite{sun2018pix3d} chair. 1 or 2 under LSTM denotes the number of LSTM units the model uses to capture primitive sequence in different order. We conduct the experiments in drop-one-out manner.}
	\label{ablation_recall}
	\begin{center}
		\resizebox{0.9\textwidth}{!}{
			\begin{tabular}{ccccccccccccccccc}
				\toprule
				& \multicolumn{3}{c}{Module} & \multicolumn{6}{c}{Thresholded Recall (${\rm TRec}^\delta$) $\uparrow$} \\

				Proposal Network & LSTM & Sem & BConv & ${\rm TRec}^{0.1}$  & ${\rm TRec}^{0.2}$ & ${\rm TRec}^{0.3}$
				& ${\rm TRec}^{0.4}$ & ${\rm TRec}^{0.5}$ & ${\rm TRec}^{0.6}$\\ 
				\midrule
				\midrule
				
				Ours Sem
				&2&\checkmark&\checkmark
				&\textbf{14.2}&\textbf{43.2}&\textbf{63.5}&\textbf{77.1}&\textbf{87.1}&\textbf{92.1}\\
				
				\midrule
				\multirow{2}{*}{} 
				
				&2&\checkmark&-
				&13.8&41.3&61.4&74.9&86.7&92.0\\
				&2&-&\checkmark
				&7.9&30.4&48.9&67.2&82.1&90.4\\
				&1&\checkmark&\checkmark
				&9.5&33.2&50.6&65.5&77.3&82.7\\
				&Faseter R-CNN&\checkmark&\checkmark
				&8.7&30.8&47.3&58.3&69.4&77.6\\
				\bottomrule

			\end{tabular}
		}
	\end{center}
	
\end{table*}

\section{Further Analysis}\label{sec:exp-analysis}

\subsection{Ablation on Proposal Network}
We also evaluate our proposal network design comparing to other variations in Table~\ref{ablation_recall}. 
Similar to ${\rm TAcc}^\delta$, we also use \emph{thresholded recall} (${\rm TRec}^\delta$ [\%]) to evaluate the quality of proposals, which is the percentage of ground truth primitives that $H(\mathcal{V}_i, \mathcal{V}_i^*) / L(\mathcal{V}_i^*) < \delta$, where for each ground truth primitive $\mathcal{V}_i^*$, $\mathcal{V}_i$ is its nearest match in predicted primitives. 
Note that each predicted primitive can only match one ground truth primitive. 
Our proposal module achieves the best ${\rm TRec}^\delta$ with all thresholds, which verifies the capability of our proposal network to capture rich semantic dependency and to attend to local image features.

\subsection{Robustness to Occlusion}
In Table~\ref{pix_splits}, we show results on different subsets of Pix3D chair according to truncation and occlusion. 
\textbf{Ours Sem} consistently outperforms 3D-PRNN and PQ-Net on all different subsets. 
In addition, our performance stays comparable in truncated and occluded subsets, which further demonstrates the robustness of our method.


%

\subsection{Failure Cases}

Our results include two typical failure cases: 1) novel object instances with uncommon 3D shapes; 2) heavily occluded objects. Two examples are shown in Fig.~\ref{fig:failure_case}. 
Both scenarios can potentially be improved with better modeling of object priors or utilizing larger datasets.


\end{document}